%% file: confpaper2.tex
\documentclass[conference]{IEEEtran}
\usepackage{blindtext, graphicx}
\usepackage{amsmath}
\usepackage{url}
\ifCLASSINFOpdf
\else
\fi
\begin{document}
%
\title{Low Cost Autonomous Navigation and Control of a Mechanically Balanced Bicycle with Dual Locomotion Mode}

\author{\IEEEauthorblockN{Ayush Pandey, Subhamoy Mahajan, Adarsh Kosta, Dhananjay Yadav, Vikas Pandey, Saurav Sahay,\\
Siddharth Jha, Shubh Agarwal, Aashay Bhise, Raushan Kumar, Aniket Bhushan, Vraj Parikh,\\ Ankit
Lohani, Saurabh Dash, Himanshu Choudhary, Rahul Kumar, Anurag Sharma, Arnab Mondal,\\
Chendika Karthik Sai, P N Vamshi}
\IEEEauthorblockA{\textbf{Indian Institute of Technology}\\
\textbf{Kharagpur, India 721302}\\
\textbf{Email: ayushpandey@iitkgp.ac.in}}
}


%


\maketitle

\begin{abstract}

On the lines of the huge and varied efforts in the field of automation
with respect to technology development and innovation of vehicles to
make them run on electric power and moreover autonomously, this
paper presents an innovation to a bicycle. A
normal daily-use bicycle was modified at low-cost such that it
runs autonomously, while maintaining its original form i.e. the
manual drive. Hence, a bicycle which could be normally driven by
any human and with a press of switch could run autonomously
according to the user’s needs has been developed. 
\end{abstract}


%
\IEEEpeerreviewmaketitle

\section{Introduction}
The paper is divided into three sections broadly, viz.
the mechanical design, the control and planning and finally the
actual implementation of the bicycle along with the limitations and
applications of the product. The point kept in mind throughout the
design process was that the full cost of the modifications made to the
bicycle should be low and the electrical and computational
complexity as well should be minimized so as to lead to robust
product. These requirements were met to a great extent as shall
be shown in the following sections.\\
Any autonomous vehicle along with all its advantages of reducing
human effort and better repeatability etc. carries some disadvantages
which are very well known and are majorly the reasons why
autonomous vehicles aren’t yet on the roads and are far from daily
usability. The biggest advantage of this autonomous bicycle, (named
i-Bike) is that it takes the best of both the worlds, the
robustness and fail-safe properties of manual vehicles along with the
low human effort, user-friendliness and advanced-innovative
tools available in an autonomous vehicle. The i-bike successfully
achieves that.\\
Added to this, during the design process of the i-Bike it was realized
that achieving a full fledged autonomous bicycle is a humongous task
(pertaining to its unstable nature) and one that requires great
efforts to achieve robustly. Hence, the identity of the bicycle wasn’t
sacrificed by making only as few modifications (reversible ones) as
possible at low-cost.\\
The motivation to build this product comes from the fact that it’s
always difficult and time consuming to retrieve one’s bicycle from
the parking stand (for eg. in a bike-share system). It is even worse for someone who has a disability
or is a differently-abled person, people who can drive a bicycle but
face a lot of difficulty in getting it back from the parking zone, which
usually turns into a disorganized area, especially in a country like
India. The i-bike solves this problem along with being user friendly
as described in the last section and has many other applications as
well. It has certain other features as well such as tracking through
GPS; wireless GSM based control and vehicle security applications.
In essence, the modifications brought in serve the purpose to a great
extent and are quite robust and feasible both in terms of its cost and
durability. The initial two sections describe the technical details of
the modifications made to the bicycle, mechanically (first section) 
and in control and motion planning (second section).

\section{Background}
The problem of autonomous stabilization of bicycles has been an area
of research for many years and a lot of methods have been proposed.
One of the most common methods of balancing a bicycle is the use of
inverted pendulum model. In this method, the balancer (inverted
pendulum) configuration is changed according to the states of motion
and position of the bicycle elements. Experimental study of balancing
an autonomous bicycle with a balancer has been reported in [2] and
[3]. Stabilizing the bicycle with additionally controlling the steering
has much better performance and has been reported in [4] and [5].
Further works have been reported which involve acrobatic maneuvers
of a bicycle using a new balancer configuration [1]. In all of the
reported experimental setups, the balancer is controlled by a motor,
thus constantly consuming energy and insensitive to abrupt changes
in the configuration of the bicycle which will require faster motors with
higher torques. However, this paper presents a method to attain the
same with ease using a mechanically implemented control
which stabilizes the roll of the system (passive stabilization). The research focuses on the
novel system design to achieve the stability mechanically and to
convert the bicycle into a drive by wire system with proper
modifications to the braking and driving. The uniqueness of the
design is that the bicycle doesn't lose its normal operation and can be
operated in a dual locomotion mode (i.e. can be switched back with ease to its normal manual operation from autonomous mode). 

\section{Mechanical Design}
The mechanical design of each and every component is such that it
only adds onto the existing structure of the bicycle and any
modification does not interfere with riding capabilities of a normal
user or rider. The volume a rider may occupy was also kept in mind
while designing each and every structure. 
\subsection{Drive Mechanism}
Driving Mechanism [6] refers to the complete mechanism designed
for the translation of the bicycle. In autonomous mode, the drive
mechanism enables it to translate both in forward and backward
directions. There are various ways in which driving can be achieved in
vehicles. Electric motors are the most commonly used actuators due
to their easy availability, simple setup and cheap nature. The i-Bike
uses electric motors as well. Power transmission can be achieved using various methods
like chain drive, shaft drive, wire rope and pulley drive. Hydraulics
can also be used as well for power transmission. But the most economical
and viable option was the chain and sprocket mechanism which was used
in the i-Bike.\\
A normal bicycle has a free wheel on the rear hub shell and there is
ample space for a motor driven chain-sprocket mechanism to be
installed for automation. The motor was placed on a base
supported on the rack of the bicycle (as shown in Figure) because it
provides a firm and rigid base reducing the need to make design
changes to the bicycle.\\
The freewheel was modified which works as a sprocket. To
provide for both autonomous and manual cycling modes, a
mechanism was provided for engaging/disengaging the motor
from chain-sprocket assembly by simply sliding and locking it. 
Disengaging/engaging mechanism was chosen such that the cost
of manufacturing is low and is easy to use. Proper SolidWorks
designs were made and tested for strains before fabricating on the bicycle. Two shafts were designed based on
lock and key mechanism mentioned. One of the shaft was attached to the motor's
shaft and another part was attached to modified freewheel supported
by two self-aligning ball bearings. The use of self-aligning ball
bearings reduces high accuracy needs which is otherwise very
important to align the motor's shaft with the modified
freewheel’s shaft. The mechanism is engaged when the two components move towards
each other and unlocked when moved away from each other. To
allow for such a movement, motor is provided with a base which slides on
the main base. The shaft attached to the motor was of hexagonal shape
and the shaft on the sprocket had its counterpart.\\
The shape of the shaft was chosen such that the engaging process is
easy and there would be less wear in the shaft. Hexagonal shape was
chosen so that cyclist could rotate the shaft by 30 degrees maximum
(~15cm forward or backward) for engaging the motor. The material
used for the shafts was mild steel and the lock and key part was flame
hardened to reduce wear and tear.

\subsection{Steering Mechanism}
Efficient steering mechanism [7],[8] is required to control the bicycle
autonomously and its design plays a very major role in position
control as it controls the direction in which the bicycle would move.To
steer the bicycle, the motor needs to overcome the torque on the stem due
to gravity and centrifugal torque, gyroscopic effects and momentum
induced torque. We calculate the required motor torque using the following equation.
\begin{equation}
Torque= \Delta_{scalar}\theta_{f}\sin(\phi) + \sigma \cos(\phi) \frac{A}{L}Mg
\label{torque}
\end{equation}
where, $\Delta_{scalar}$ is scalar value of trial vector, $\theta_{f}$ is the lean axis of the frame, $\phi$ is the steering axis angle, $\sigma$ is the turn angle of the handle bar, $L$ is the distance between the wheel hubs and $A$ is the horizontal distance between the rear wheel hub and the centre of mass. The required torque using Eq.(\ref{torque}) comes out to be 2.7 Nm.\\
A servo motor matching required torque was used to steer the
bicycle which was mounted parallel to the head tube. A chain sprocket
mechanism was used for power transmission. A sprocket freely moves
below the headset and can be tightened using a lever welded to
headset. On tightening the sprocket motor is engaged with the stem of the bicycle due to high friction and to disengage the motor sprocket is
loosened. The motor was mounted using clamp attached to top tube and
parallel to the head tube. 
\subsection{Braking Mechanism}
Three mono pivot breaks were used for braking, two of which
were at their normal position which were manually operated and the third
was attached on the front wheel opposite to the normal position of
manual brake.\\
Autonomous braking was controlled by a DC motor
which rotates a disk on which a wire gets wound up. The motor when
switched on pushes the brakes pads towards the wheel, but
prolonging the same may damage the motor.
\subsection{Balancing Mechanism}
Usually bicycles are balanced with electronic feedback control; in
this bicycle the aim was to reduce the already complex system by using
mechanical feedback control. There can be many such control
systems; we chose to use a torsional spring and damper system for
the same.\\
A pair of balancing wheels was used which remain in contact with the
ground even when bicycle is upright. They were attached on
the axle of rear wheel hub. A normal set of training wheels would not
allow the bicycle to lean while taking turns. A balancing wheel
includes a torsional spring on which a freely rotating wheel was
mounted. 

\begin{figure}[htbp!]
			  \centering
			 \tiny{	
			\resizebox{5cm}{!}{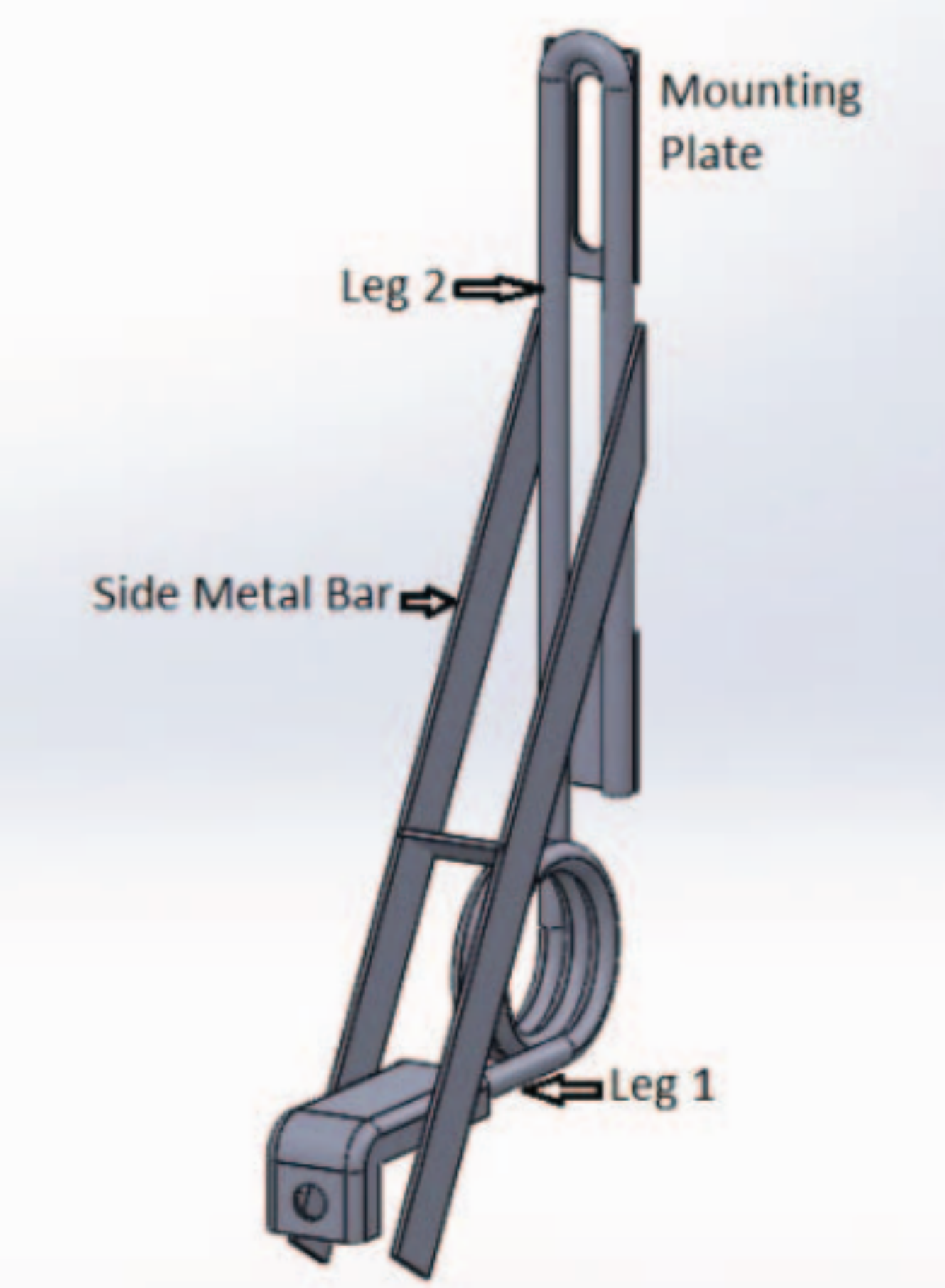}}
			  \caption{Spring balancer model}
			 \label{balancing}
		\end{figure}	
\textbf{Spring Design: }Both springs were helical torsion spring with right
handed helical loop for left spring and left handed helical loop for
right spring. If there is no restriction, leg 1 of spring acts as a cantilever beam and vibrates about vertical axis. The two side metal
bars restrict the vibration of the spring but allow winding of spring. A
plate with slot was welded on other leg of spring. The rear wheel hub
bolt come through this slot and whole assembly was tightened by using a nut.
The slot on plate allows adjusting height of balancing wheel.
Calculation of Spring Constant: Considering space constraints the
mean coil diameter of spring was kept at 5 cm. The radius of rear wheel of
bicycle was 30cm and of trainer wheel was 6cm. Therefore, length of leg 1
$(l_{1})$ is given by:
\begin{equation}
l_{1} = 30-6-2.5 = 21.5cm.
\end{equation}
If the bicycle tilts in one direction while turning or due to some
obstacle; it should be able to regain vertical position all by itself.
Considering a situation where the bicycle is tilted and makes an angle $\theta$ with
vertical. Therefore, spring also winds by an angle of $\theta$. 

\begin{figure}[htbp!]
			  \centering
			 \tiny{	
			\resizebox{7cm}{!}{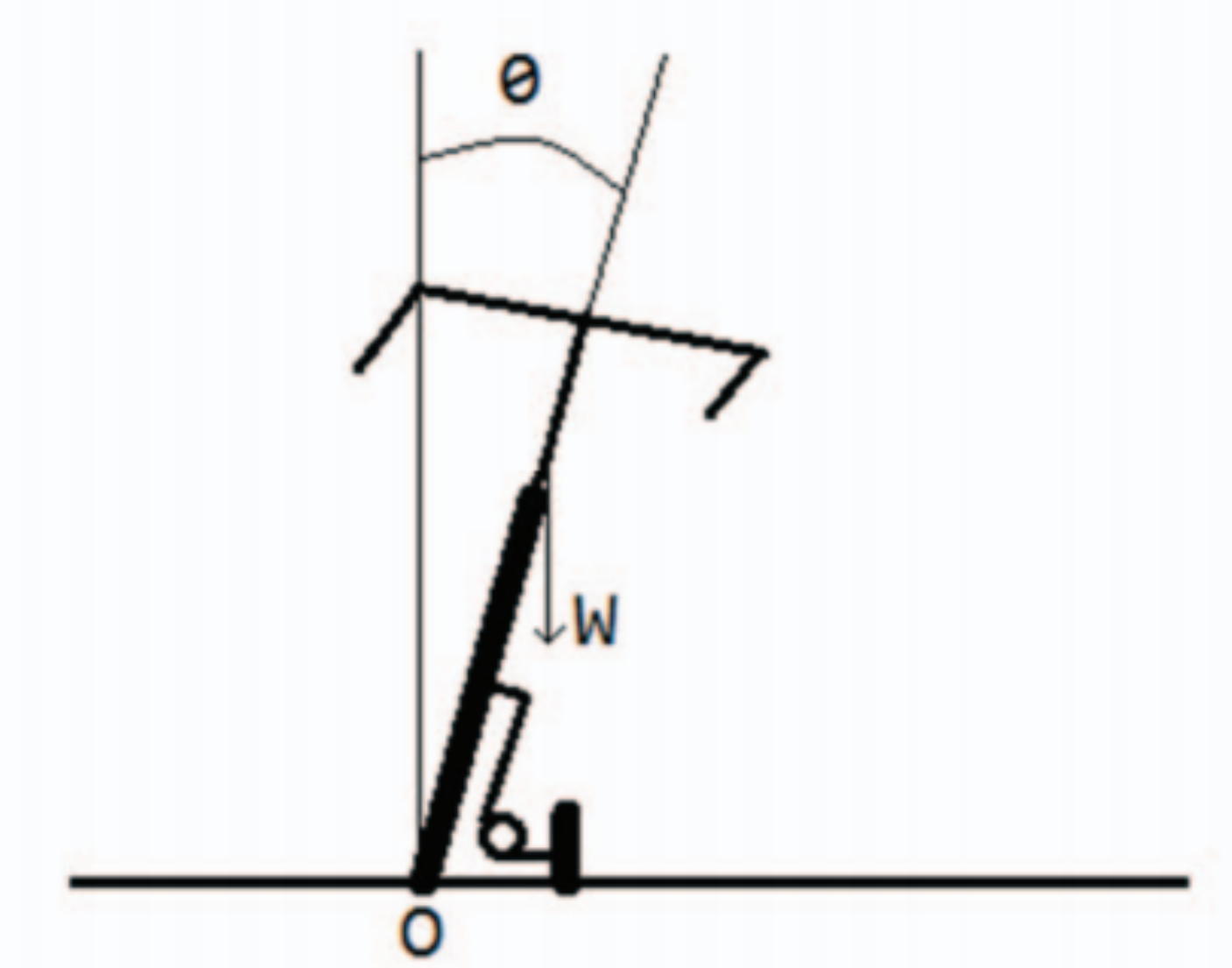}}
			  \caption{Free body diagram for torque calculation}
			 \label{balancing}
		\end{figure}	
Torque due to weight of bicycle about point O is given by the following equation.
\begin{equation}
T_{w} = mgh\sin(\theta)
\end{equation}
where $m$ is the mass of the bicycle without one balancing wheel (25 kg in our case), $g$ is acceleration due to gravity and $h$ is the height of the center of mass of the bicycle from ground in the vertical position (which is $0.6m$ in our case).\\
The torque due to spring about point O is then given by
\begin{equation}
T_{s} = \frac{kșR}{l_{1}}
\end{equation}
where $k$ is the spring constant (in Nm/radian), $R$ is the radius of the rear wheel (equal to $0.3m$ in our case) and $l_{1}$ is the length of leg 1 ($0.215m$ for our bicycle design).\\
For the bicycle to regain vertical position, there must be a net
counterclockwise torque about point O for every value of $\theta$ greater
than $0$.
\begin{figure}[htbp!]
			  \centering
			 \tiny{	
			\resizebox{7cm}{!}{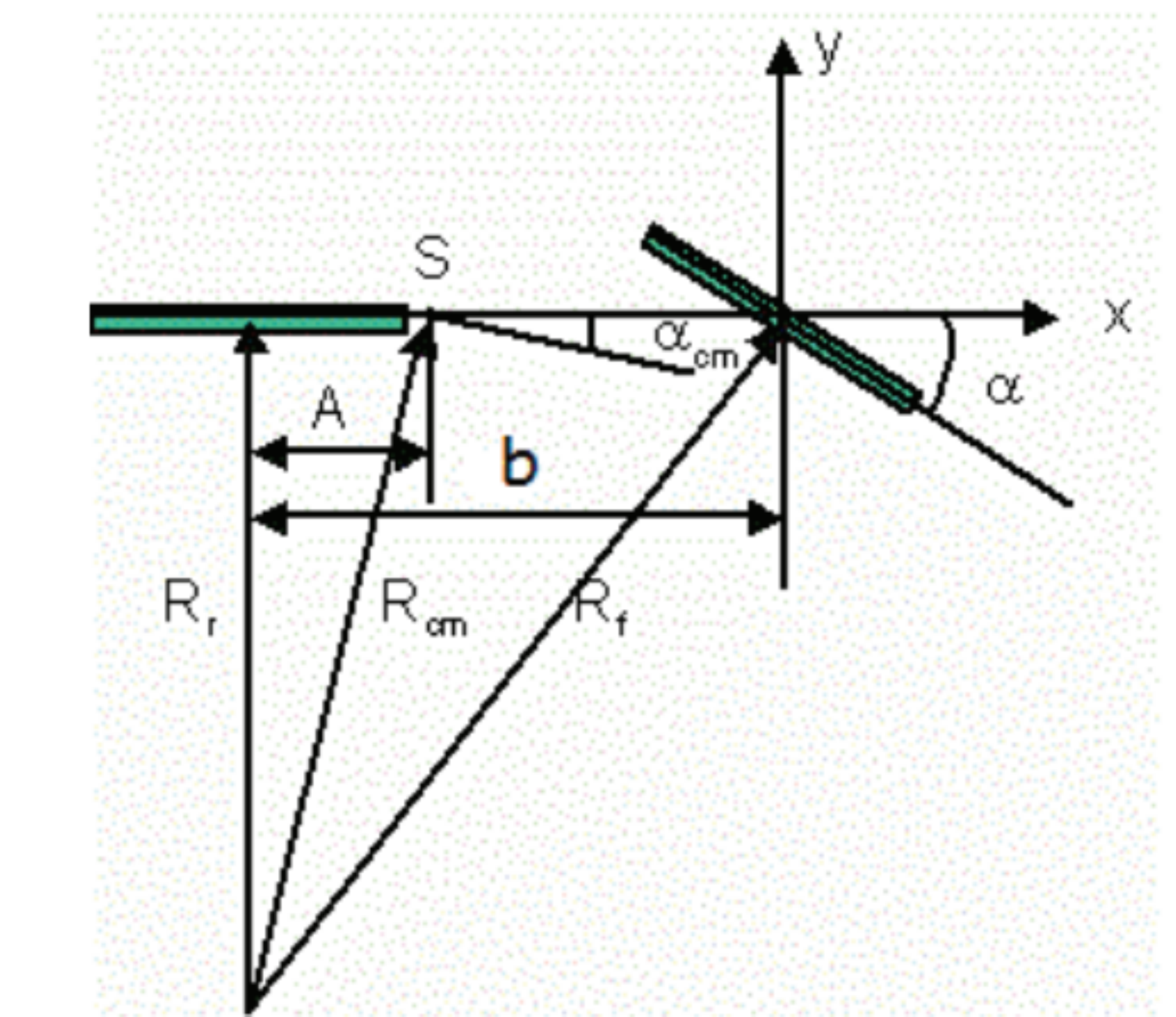}}
			  \caption{Balancing Wheel Analysis, where $b$ is the wheel base (1m); $A$ is the horizontal distance of the center of mass of the bicycle from the rear wheel axle (.4m); $Į$ is steering angle and $Rcm$ is the radius of the turn of the center of mass of bicycle}
			 \label{geometry}
		\end{figure}	
\begin{align}
T_{s} - T_{w} &> 0\\
T_{s} &> T_{w}\\
\frac{k \theta R}{l_{1}} &> mgh\sin(\theta)\\
\frac{kR}{mghl_{1}} &> \frac{sin(\theta)}{\theta}\\
\intertext{As $\theta \rightarrow 0$ we have, $\frac{\sin(\theta)}{\theta} \rightarrow 1$. So, we can write}
\frac{kR}{mghl_{1}} &> 1\\
\intertext{We get the value of $k$ on substitution}
k &> 117.7 Nm/rad\\
\intertext{Now from Fig(\ref{geometry}) we can write}
R_{cm} &= \sqrt{ A^{2} + \left(\frac{b}{tan(\alpha)}\right)^{2} }\\
\intertext{Using the above relation we conclude that the maximum steering angle should be $30^{\circ}$. Hence, the smallest possible value of $r$ would be $1.78m$. Now, we can write the torque of the centrifugal force about point O, while taking a turn as follows}
T_{c} &= \frac{mv^{2}h\cos(\theta)}{r}\\
\intertext{where $m$ is the mass of the bicycle, $v$ is the velocity of the bicycle and $r$ is the turning radius. Although the maximum possible velocity of bicycle is $5m/s$, while turning it would be brought down to $2m/s$ to reduce the centrifugal force. Hence, to obtain the maximum value of $T_{c}$ we would use the maximum value of velocity($2m/s$) and the minimum turning radius to be equal to $1.78m$. At this condition, the bicycle would not tilt by more than $10^{\circ}$. Therefore, at $\theta=10^{\circ}$ the net torque about point O would be equal to zero.}
T_{s} – T_{w} – T_{c} &= 0\\
T_{s} &= T_{w} + T_{c}\\
\frac{kșR}{l_{1}} &= mgh\sin(\theta) + \frac{mv^{2}h\cos(\theta)}{r}\\
\intertext{Substituting for the parameters, we get $k = 4.9$ Nm/degree, using}
k &= \frac{d^{4}E\pi}{64 \times 180 DN_{a}}\\
\intertext{where $d$ is the diameter of the spring wire, $E$ is the modulus of rigidity of stainless steel (180GPa), $D$ is the mean coil diameter (5cm) and $N_{a}$ is the equivalent number of active turns}
N_{a} &= N_{b} + \frac{l_{1} + l_{2}}{3\pi D}
\end{align}
where $N_{b}$ is the number of body turns (2.25). Substituting the values and equating to $k=4.9$ Nm/deg, we get $d=1.47$cm.
\\\\
\textbf{Calculation of maximum safe compression of spring}\\
Minimum tensile strength of spring $S_{ut} = \frac{A}{dm}$. For stainless steel wire, $A= 2911 MPa-mm^{m}$ where $m=.478$. Torsional yield strength $S_{y} = .61 S_{ut} = 491.4 MPa$. Bending stress for round wire torsion spring is given by the following equation
\begin{align}
\sigma &= \frac{32K_{i}M}{\pi d^{3}}\\
\intertext{Where $K_{i}$ is the bending stress correction factor and $M$ is the moment of the force acting on the spring}
K_{i} &= \frac{4c^{2} - c - 1}{4c(c-1}\\
\intertext{where $c$ is the spring index, which is equal to $D/d$. At maximum compression $\sigma = S_{y}$.}
M &= \frac{\pi d^{3} S_{y}}{32K_{i}} = 119.7 Nm\\
\intertext{Therefore, the maximum safe compression angle is given by} 
\theta &= \frac{M}{k} = 24.4^{\circ}\\
\end{align}
which is greater than $10^{\circ}$\\
A shorter length of leg 2 will increase compressive stress on trainer
wheel which was made of plastic but longer length will increase its
probability of hitting obstacles. So to optimize the length of leg 2, a
stress analysis of trainer wheel was performed in SolidWorks. 
\begin{figure}[htbp!]
			  \centering
			 \tiny{	
			\resizebox{7cm}{!}{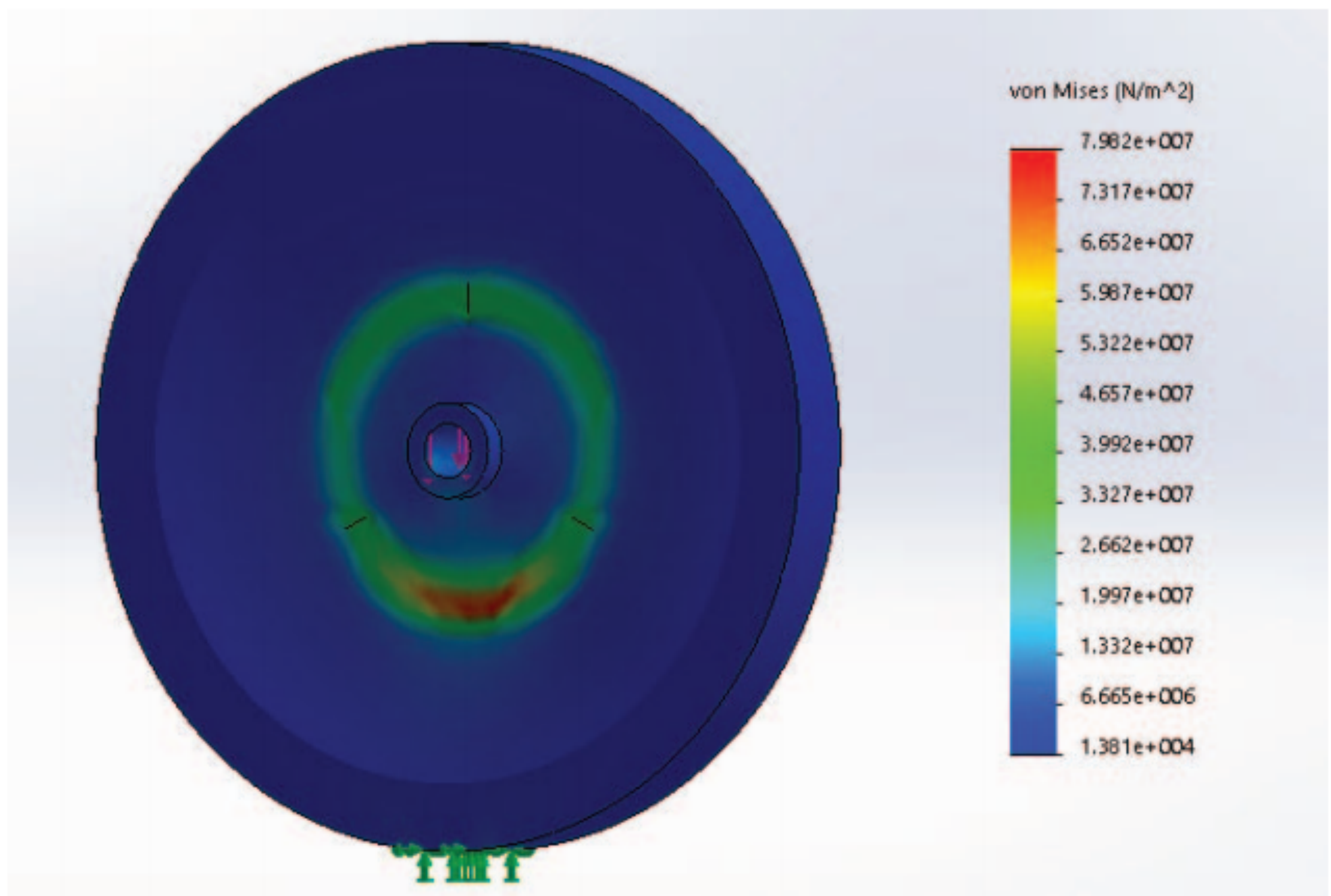}}
			  \caption{Stress analysis of the trainer wheel}
			 \label{stress}
		\end{figure}	
The moment of normal force acted by ground on trainer wheel
balances the moment of spring. So, maximum normal force will
occur when spring is compressed at maximum angle, which is $10^{\circ}$
\begin{equation}
N_{max} = \frac{M_{max}}{l_{2}}
\end{equation}
where $M_{max}$ is the maximum value of moment (641.5 N) and $l_{2}$ is the length of leg 2.\\
Trainer wheel was modeled in SolidWorks and High Density
Polyethylene (HDPE) material was assigned. Stress analysis was
performed at different Nmax by increasing values of $_{2}$ until sufficient
factory of safety was obtained. At $_{2} = .11m$, Factor of Safety of 2
was obtained. Hence length of leg 2 was kept at $.11m$.

\section{Control and Motion Planning}
The mechanical modifications made in the bicycle need to be
actuated properly and the required peripherals were added to
complete the automation process. The complete process is divided
into various parts described in detail as subsections. 
\subsection{Drive Control System}
The translation motion of the i-Bike, both forward and backward, and
hence the velocity control is achieved using a feedback control
system. The actuators used in the control system were two high torque
DC motors [9]. The sensor giving the velocity feedback in the form
of counts per revolution is the rotary encoders placed around the
motor shaft. The control was achieved using a PID controller algorithm
[10] implemented on an electronics chip.\\
The steer motor was coupled using a sprocket and chain mechanism
directly to the main handle of the bicycle and the drive motor was
coupled using a similar mechanism to an extension to the back wheel
of the bicycle. As mentioned earlier, the bicycle can move using human
power and also can be motor driven. So for achieving this, an
engaging-disengaging mechanism was added to the drive motor for
control method chosen. 

\begin{figure}[htbp!]
			  \centering
			 \tiny{	
			\resizebox{7cm}{!}{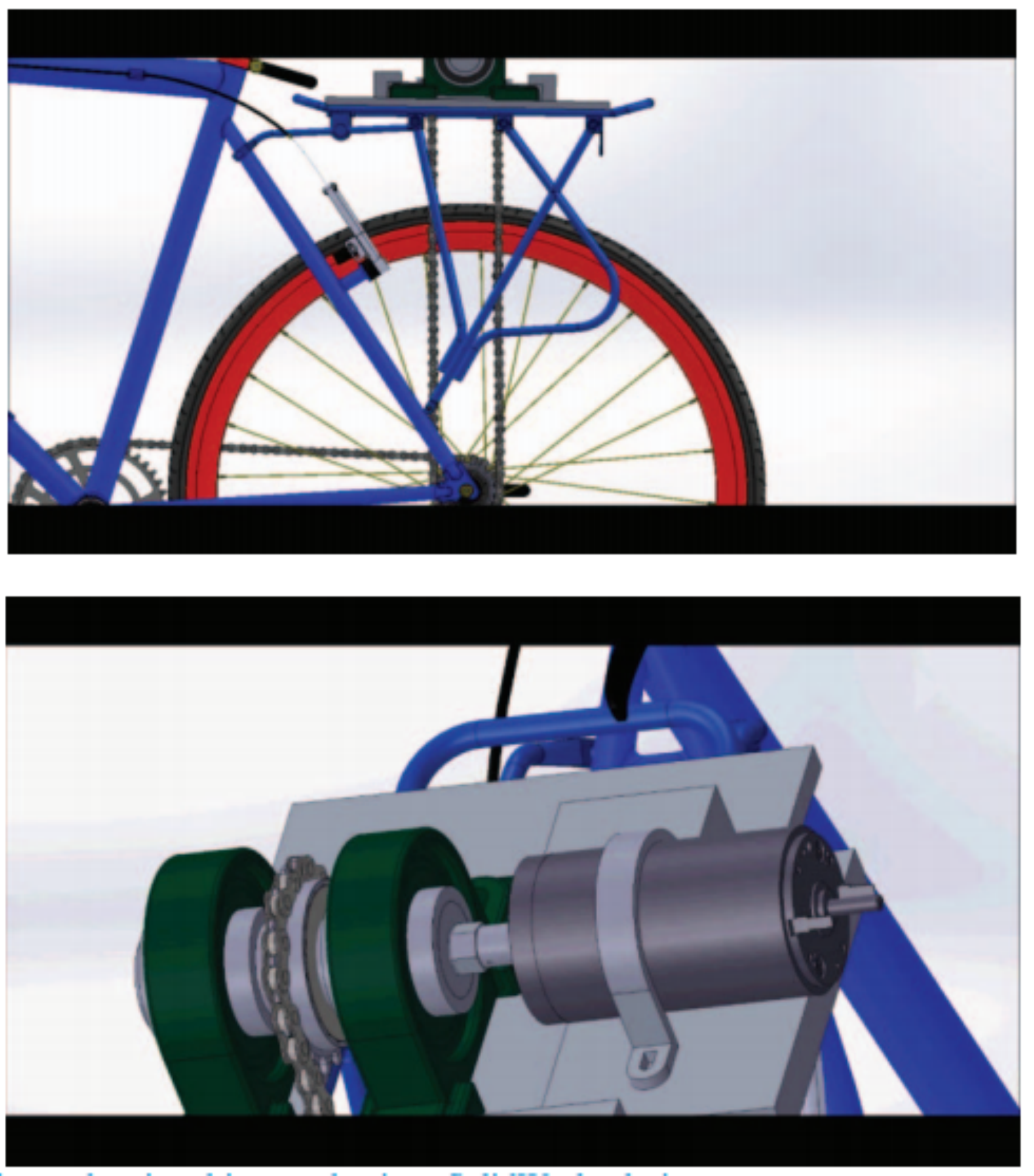}}
			  \caption{Drive control mechanism}
			 \label{drive}
		\end{figure}	
The motor was driven using a pulse given by the microcontroller, the
Arduino having ATmega 2560 chip on which the control algorithm was
implemented. The frequency of the PWM decides the speed to which
the motor is sped.\\
The motor requires quite a high amount of current at full load (7 A)
and this was achieved using lightweight but high capacity Lithium
Polymer (LiPo) batteries of 2200mAH. The motors were successfully
able to do both operations, given their high amount of advertised
torque (120 Kg-cm). The battery lasts through about 2 hours on a
single charge. This may be improved further by using motors with
lower current rating.\\
Tests were performed on the mechanisms for testing for any kinds of
slipping due to mechanical fault and results were satisfactory in case
of the drive motor, but the steer motor encountered a lot of slipping in
a particular direction, which was corrected by an LED-LDR pair (an
Opto-Coupler).

\begin{figure}[htbp!]
			  \centering
			 \tiny{	
			\resizebox{7cm}{!}{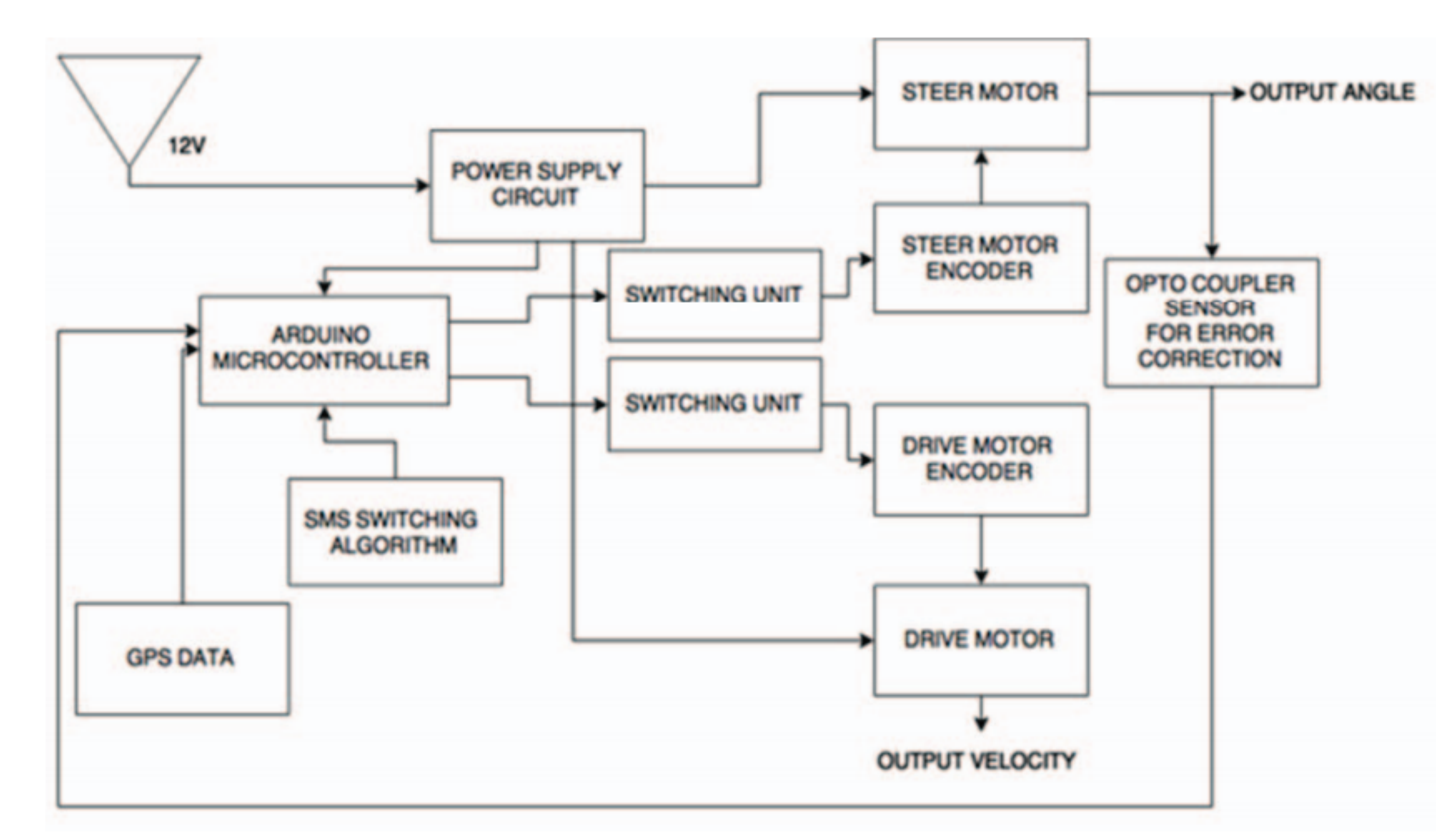}}
			  \caption{Flow chart depicting control system schematic}
			 \label{flow}
		\end{figure}
For the localization of the bicycle, a basic form of odometry [11],
[12] was also implemented using simple geometric formulae and
using data of encoders on the motor, which were correct to an error of
5-7
can be perfected by using encoders on the wheel rim and calculating
the distance using real world data, incorporating all kinds of
mechanical errors. This would make the path planning code work
more accurately. For now, The i-Bike relies on the motor encoder
assuming that the control mechanisms inside work perfectly. In future
implementations, external control systems might be used to
incorporate errors. 

\subsection{Localization}
Magnetometers were used on top of the bicycle for yaw measurement
during the motion. Since magnetometer provides the angle with
respect to the geometric north, Yaw was measured every time assuming
linear variation of magnetometer angle change with the yaw for a
very small angle.\\
For a more precise localization of the bicycle [11], along with
Yaw measurement for orientation, wheel odometry for the drive
motor was implemented. These calculations lead to the position
of the bicycle known to the controller at every instant with a fair deal
of accuracy. The counts from the encoders of the drive and steer
motor help us to calculate angle and distance travelled by the bicycle
in discrete time steps using the model that it rotates in a circle for
every small time interval. 

\subsection{Steer Angle Measurement and Control}
A dc motor was used to actuate the steering control system in the
bicycle. The complete control system for the steering was achieved in
a complex way incorporating various innovations. This was needed
since the velocity control of the steer drive is not sufficient in itself for the accurate steering control. The main reason for inaccuracies is
slippage between the motor axle and sprocket in the mechanical
design. Hence, the need for a closed loop control on top of the
internal encoder based velocity feedback control was realized using a
potentiometer fixed on the drive using an L-clamp [13]. 

\begin{figure}[htbp!]
			  \centering
			 \tiny{	
			\resizebox{7cm}{!}{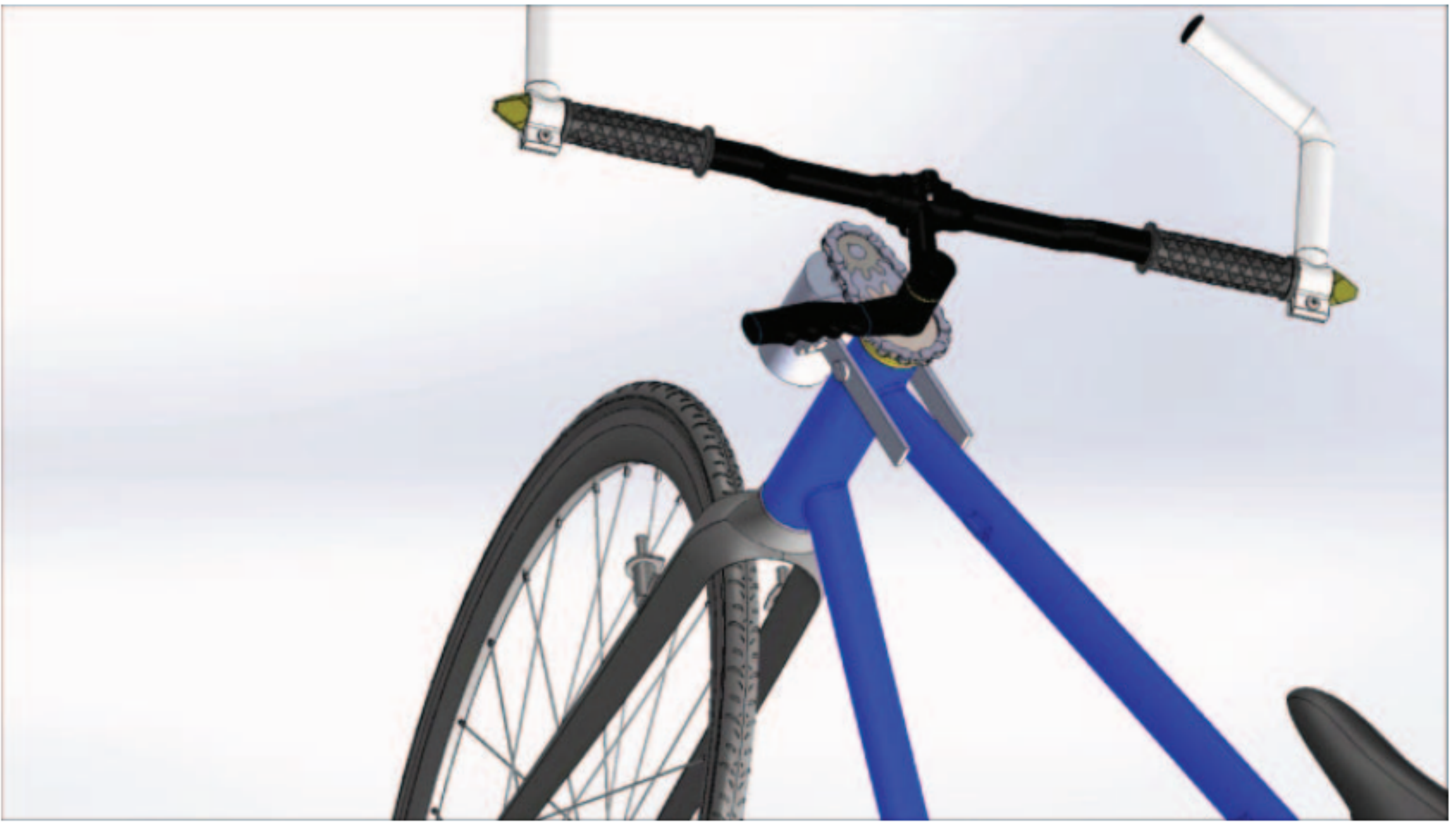}}
			  \caption{Steering mechanism design}
			 \label{flow}
		\end{figure}
During testing, the control system was further improved upon since
the central position of the bicycle was still in doubt and incorrect
results were being obtained. Hence, a custom-made encoder was
designed and placed on the mudguard of the bicycle. The encoder
was realized in the form of an IR sensor, called the MOC sensor. A
flap was designed which was placed judiciously such that every time
the wheel crossed the central position a signal was received and we
could eliminate all the errors accumulated in that run (and hence
nullify all such errors). This lead to a very accurate steer angle
control mechanism. \\
The steering microcontroller receives the data from the drive
microcontroller which acts as a path planner for the bicycle, giving it
the steering angle to rotate, after calculations and estimations
suggested by the Obstacle Detection module. \\
On encountering an obstacle, the obstacle avoidance module would be
activated instantly (prioritized over others) and its data would be overridden
on the steering control so that the safety is maintained as a first
priority. In future, there is scope to make the algorithm for obstacle
avoidance and planning more robust for better practical application. 

\begin{figure}[htbp!]
			  \centering
			 \tiny{	
			\resizebox{7cm}{!}{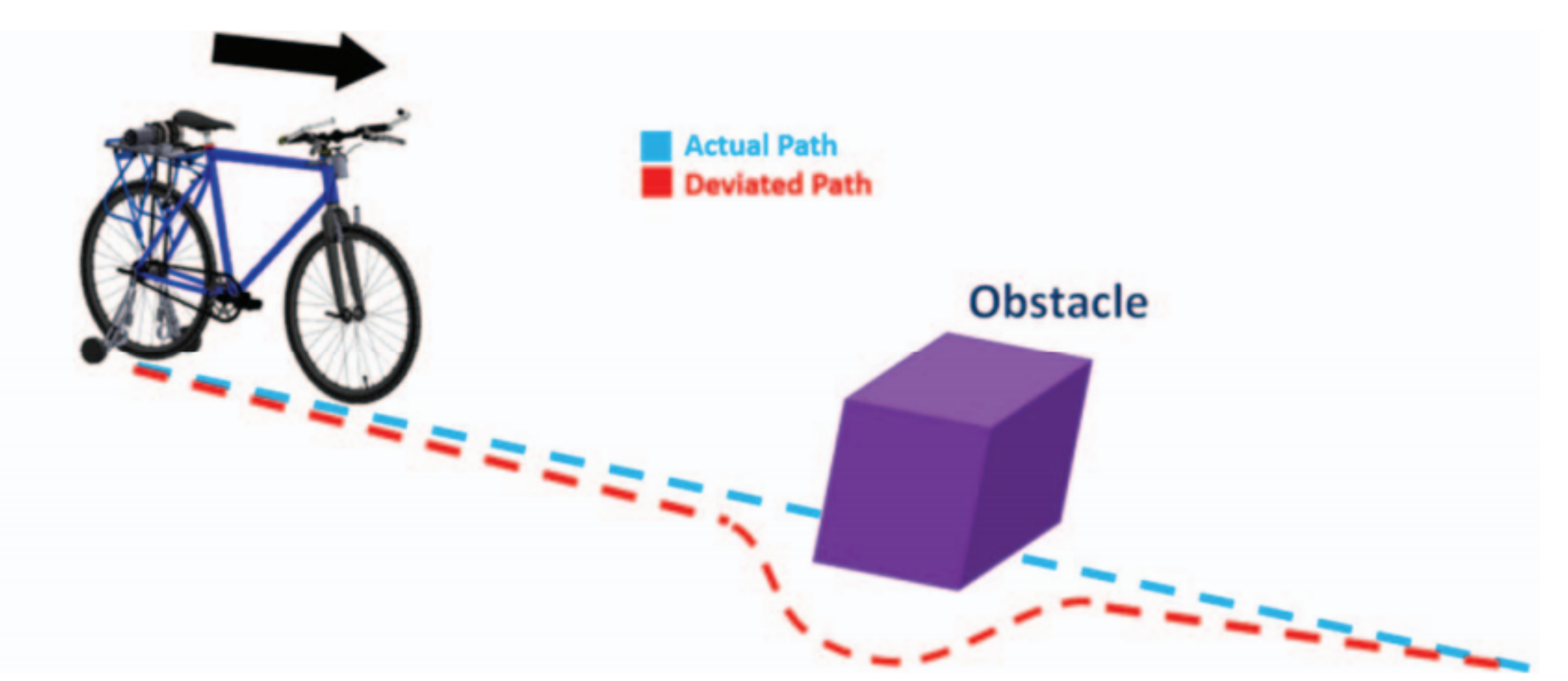}}
			  \caption{Motion planning algorithm}
			 \label{planning}
		\end{figure}
		
		\begin{figure}[htbp!]
			  \centering
			 \tiny{	
			\resizebox{7cm}{!}{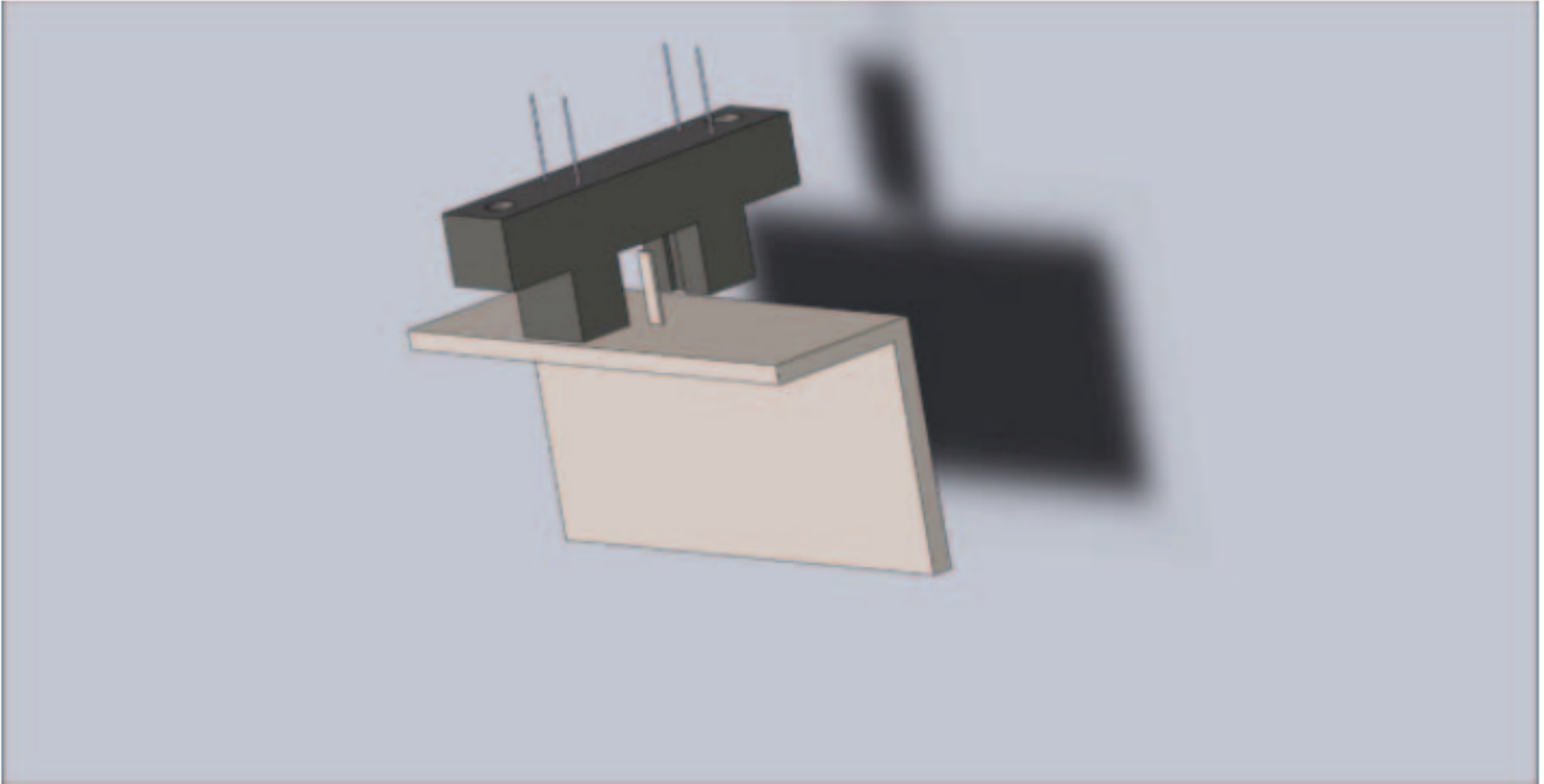}}
			  \caption{Encoder working mechanism to reduce errors in steering control}
			 \label{encoder}
		\end{figure}
\subsection{Obstacle Detection and Avoidance}
Two ultrasonic SONARs with a range of 2-400 cm were used for
obstacle detection in which time of flight of the ultrasonic wave is
determined and hence obstacle distance and angle (measured because
of the position of the sensor). \\
The i-Bike uses 2 SONARs inclined at an angle of 10 degrees with
vertical in same horizontal plane. The SONARs detect obstacles on
either side of the bicycle with overlapping cones for also detecting
obstacles in front of bicycle, covering a total obstacle detection cone
of 45 degrees. In this way, obstacle detection with a considerable
field of view was achieved. After the detection of an obstacle, the
control was shifted to the obstacle avoidance module which was
implemented on the microcontroller driving the bicycle.

\subsection{Autonomous Braking}

A low rpm dc motor with appropriate torque (to pull the brake wire) was used as the braking actuator. The braking wire was attached to the
axis of the dc motor and the other end was attached to the bicycle rear
wheel braking mechanism used in normal bicycles. A proper motor
angle range was chosen to brake the front wheel to make sure that the
braking shoes maintain a constant range of braking force required to
stop the bicycle. Currently, the system implements braking as an
open loop system, improvements could be made upon this design to
make the braking closed-loop by detecting and choosing when to
brake. Additionally, along with this braking system the drive motor
on the back was capable on itself to slow the bicycle down (or even
move it backwards!). Though, since currently the bicycle
demonstrates very low speeds such as 5m/s or even less, braking is
not much of a concern. But, the arrangement in place ensures for the
autonomous braking whenever needed.

\subsection{User-friendly Interface and Other Features}
Autonomous Navigation in the bicycle was achieved by localizing it
with respect to its initial position using global positioning system (GPS) and then it could be controlled using a SMS (received from the user)
which could either contain destination GPS co-ordinates or relative
position of the destination with respect to its initial position. This has
been discussed in [1]. SMS communication was used to communicate
between user and bicycle making it considerably simple and user
friendly. [2] explains more in detail on this fact. \\
To facilitate this SMS based system in the bicycle, GSM SIM 900
module was interfaced with the controller to receive/send messages
regarding the target location which bicycle needs to achieve or any
other control actions. The current location of the bicycle would be
determined by the GPS module.

\subsection{Online Tracking of the Bicycle}
This significant feature in the bicycle was achieved using GPRS
technology available in the GSM SIM 900 module. A GSM modem
containing a SIM card of any valid service provider was used to
send the position (latitude and longitude) of the vehicle from a
remote place over the internet. GPRS was activated on this module
using the HTTP protocol which sends the current GPS
coordinates to the web server which was created using a Javascript.
This feature adds to the safety of the bicycle.
\subsection{Data Processing and Acquisition}
In this bicycle system the data processing was done in a
distributive manner on two 8 bit 16 MHz Atmega2560 microcontrollers. The drive microcontroller plans the motion for the bicycle using the
algorithm fed to it and executes appropriate commands to the
actuators. Along with this, the drive microcontroller also receives
the GPS coordinates, SMS and call commands through the GSM Modem
and controls the drive motor. The required steer angle for reaching
the destination point using the path planning algorithm is sent to the
steer microcontroller using UART communication.
\subsection{Power Distribution and Management}
\begin{figure}[htbp!]
			  \centering
			 \tiny{	
			\resizebox{8cm}{!}{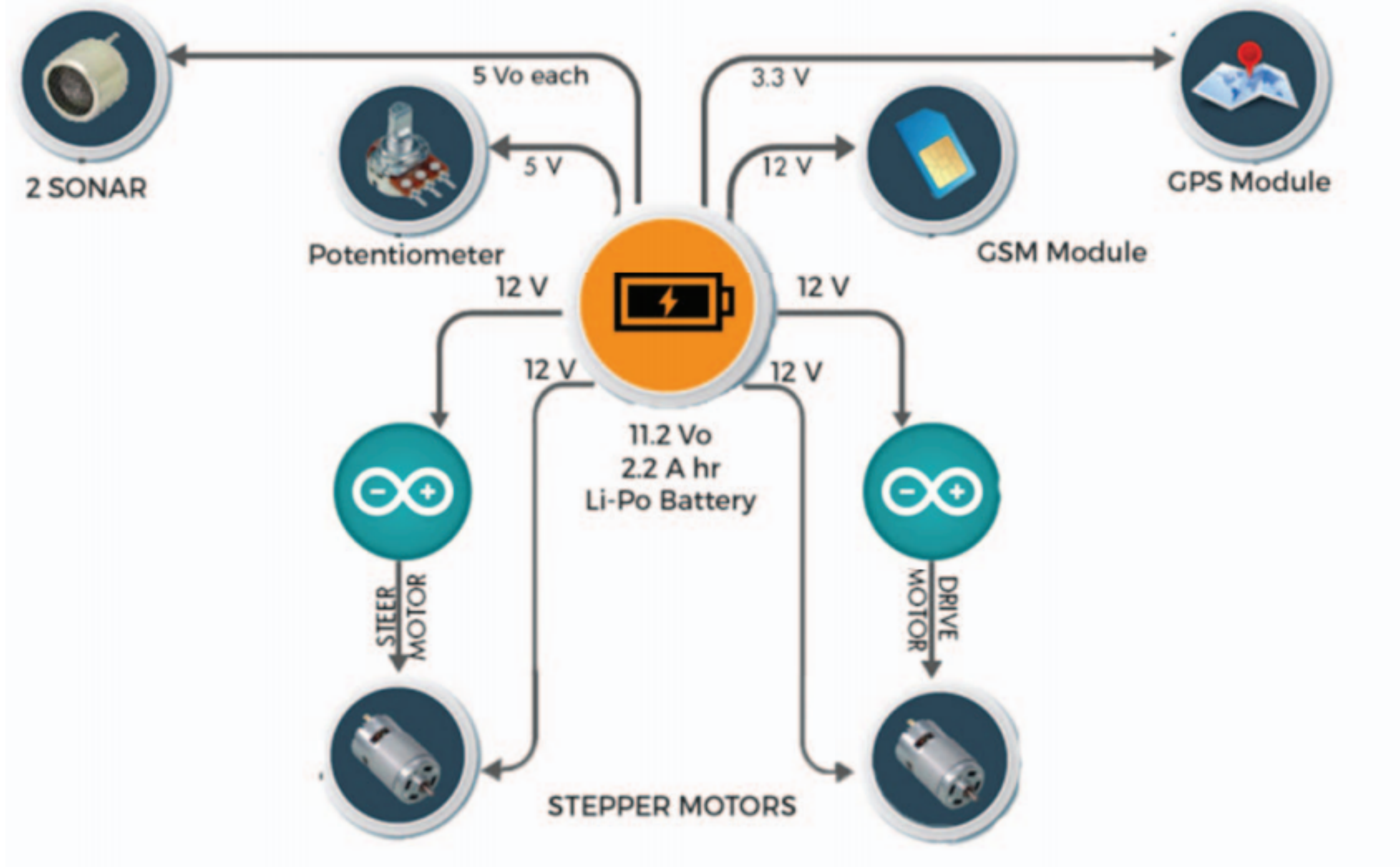}}
			  \caption{Power management and electronic architecture}
			 \label{power}
		\end{figure}

Proper voltage regulator circuits were in place to provide different
voltages to different components. The same is clear from the
electronic architecture as shown in Fig.(\ref{power}).

Each of the 12V DC servo motors draws a current of 800mA under
no load condition and 7.5A current under the maximum speed
condition. The i-Bike, being autonomous, uses various electronic
components for its operation. So, a proper power distribution system
was designed by taking into account the power ratings of all the
electronic components used. The whole system of the i-Bike was
powered using a single 11.7 V 2.2 AH Li-polymer battery. In order to
supply power to all kinds of components which differ in their voltage
and power ratings proper voltage regulator circuits were designed. \\
The motor used in the model was a 12 volt and 7.5 amps driven DC motor. The 11.7 V, 2200 mAH LiPo battery [14]
was close to the voltage rating
and with a discharge current of $30 \times 2200 mAH = 66 Amp$, it was a perfect choice. Li-Po battery
was preferred because of its light weight, high discharge rate and
relatively good capacity. This battery also supports other components
in the bike for power requirements, thus keeping a margin in Amps
available and reducing the stress on the battery and
increasing its life. Two or three mid to large capacity
lithium batteries could easily fit on one i-bike, giving potential ranges
of 50 miles (80 km) or more. \\
Well designed and soldered PCBs were used for all kinds of power
distribution and signal conditioning circuits, further adding to the
robustness of the vehicle. The table below summarizes the complete
requirements. 

\begin{figure}[htbp!]
			  \centering
			 \tiny{	
			\resizebox{8cm}{!}{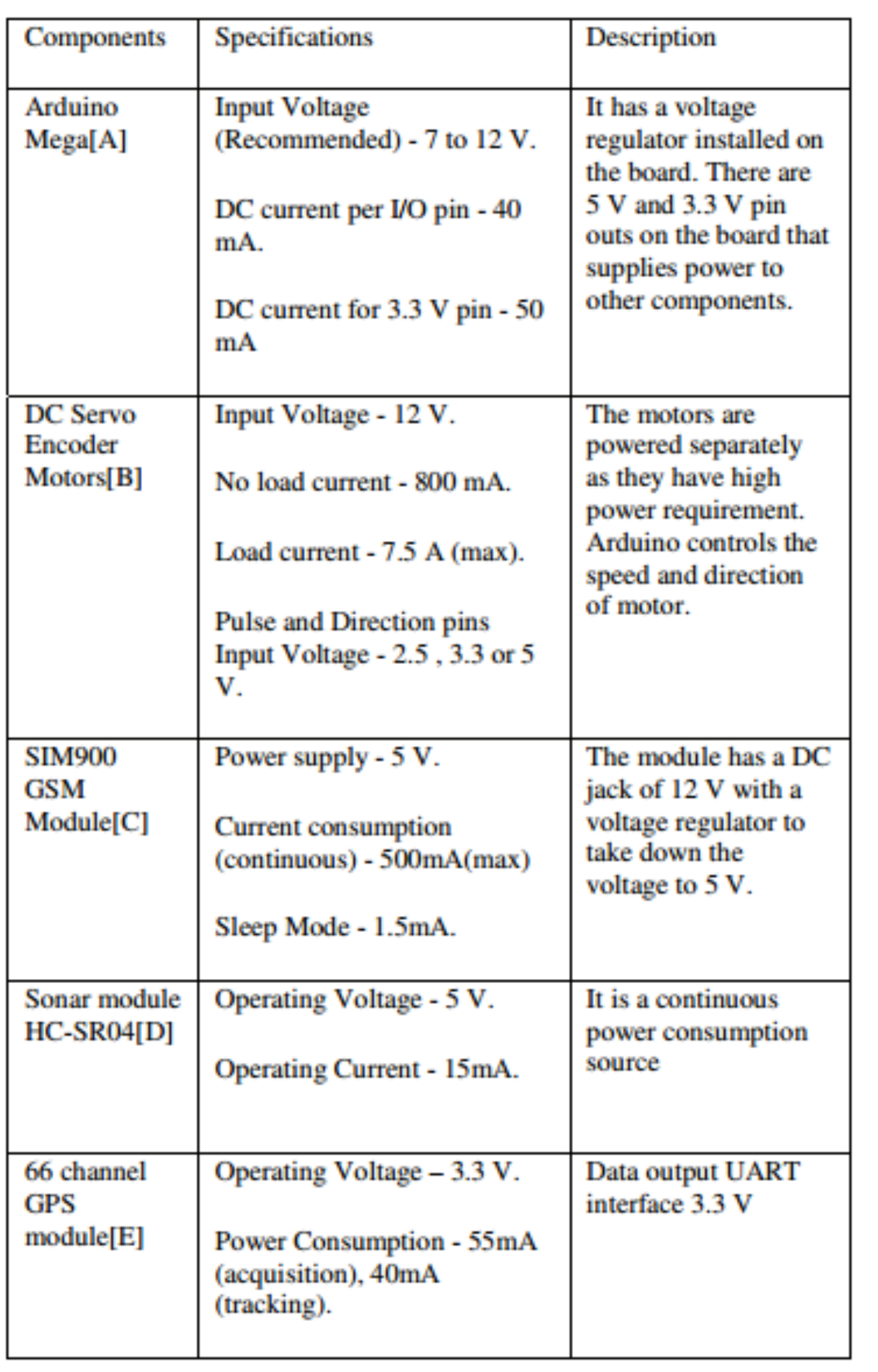}}
			  \caption{Componenet specifications and power ratings}
			 \label{table}
		\end{figure}
\section{Testing and Results}
\begin{figure}[htbp!]
			  \centering
			 \tiny{	
			\resizebox{9cm}{!}{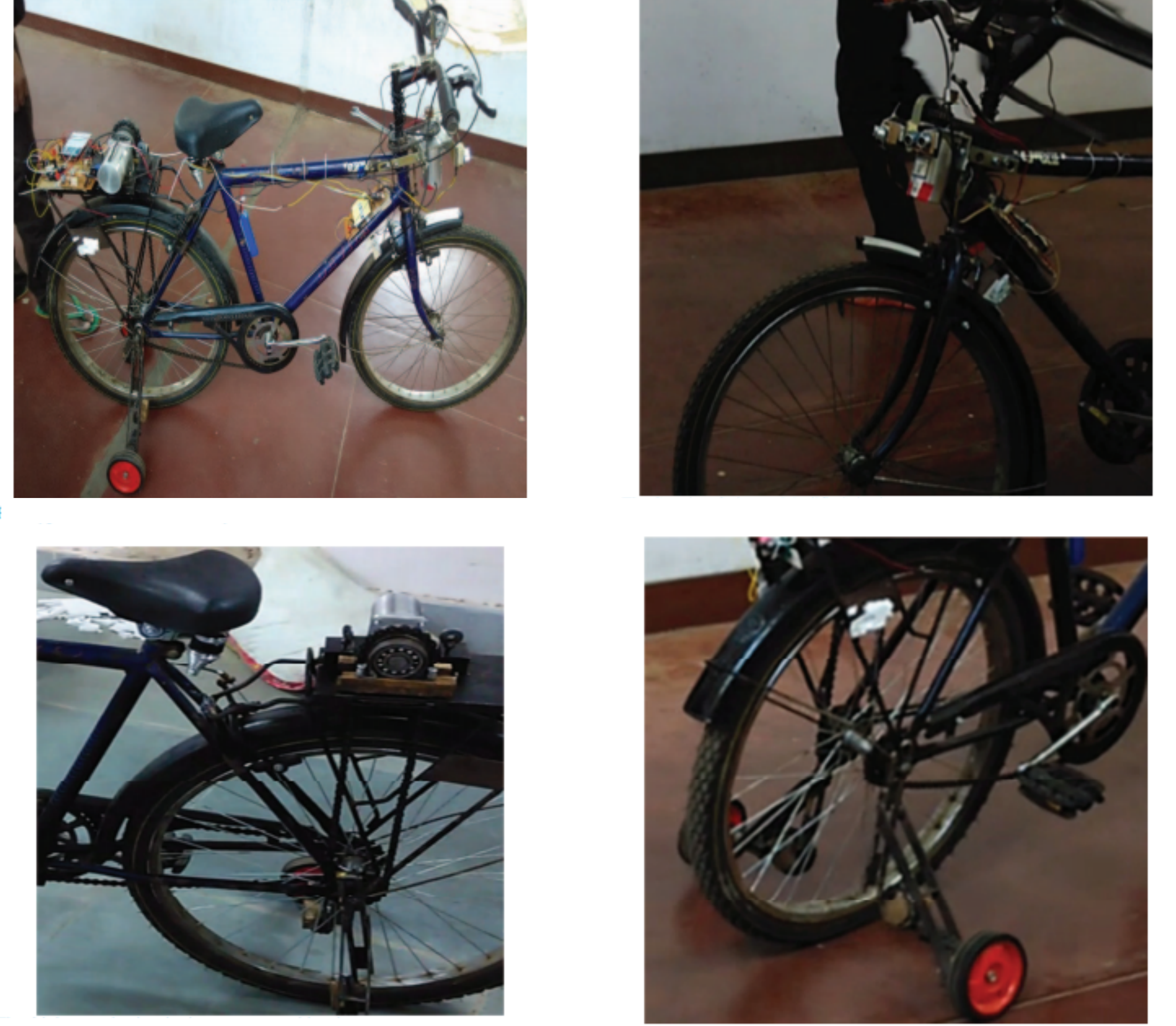}}
			  \caption{Testing and Results for the working prototype of i-Bike}
			 \label{i-Bike}
		\end{figure}
\subsection{Application areas}
There are many applications of a smart and autonomous bicycle in
today's world. It will be a boon for people who are differently-abled,
people who are visually impaired or those who have difficulty in
localization can use this vehicle to travel across a crowded city. With
the autonomous ability of the bicycle comes the freedom for multiple
people to use the same vehicle for transit. The bicycle can to rode to a
place and then be called back to be rode to a different place, just by
using mobile SMS (while maintain the security all along). The
product can also be used to deliver courier or food to customers as it
is capable to traversing through narrow and crowded lanes. With the
ability to travel through narrow roads, it would be the ideal vehicle
for Street view mapping. The painful task of street view mapping can
be done by a smart and autonomous bicycle. It could also be used for
real-time traffic monitoring as it wouldn’t be adding to the traffic. A
smart bicycle would also ease the pain of properly and systematically
parking a bicycle as one wouldn’t need to scout manually for an
available parking space. As per need, the bicycle could be called to
one’s location instead of manually walking to the parking zone. In
this modern era, we have huge factories with complex equipments; an
autonomous bicycle would be the ideal mode of transport between any
two points in the factory as it can then return to its designated space
after assisting technicians move. 
\subsection{Future prospects}
The algorithm used by the bicycle to achieve its autonomy can
be used in ally form of autonomous vehicle especially autonomous
Electric scooters. The engaging mechanism used for dual mode can
be used in other vehicles where such need might arise. The MOC
circuit used to ensure resetting of the steering even in the event of a
mechanical offset can be used in other autonomous vehicles to ensure
that the vehicle travels in a rectilinear fashion. 
\section{Limitations and Conclusions}
Current experiments with the prototype have shown obstacle
avoidance capability with sonar at slow speeds, but further work must
be done to avoid obstacles at higher speeds. The design of the spring
limits the maximum speed during a turn and the maximum tilt angle
which needs further optimizations. The design is also suitable for flat
terrain with minor disturbances but some fluctuations still exist. For
precise control of the bicycle the backlash error in the steering and
rear wheel needs to be reduced and along with that the processing
power could be increased.\\
The bicycle achieved what it set-out to, initially. It demonstrated
great robustness as any user can sit on it to drive it anywhere and then
the tests have shown that one SMS sent to the bicycle activates the
autonomous mode and the bicycle avoids all obstacles in its way,
giving the user a track of its position online, reaches the final
destination.



%


\section*{Acknowledgment}
The authors would like to thank the management committee
of the Lal Bahadur Shastri Hall of Residence, IIT Kharagpur , along
with the respected warden , Professor D. K. Maity for his kind
assistance and encouragement. We would also like to thank our
Institute authorities, the Gymkhana body (session 2014-15) and the
Gymkhana President, IIT Kharagpur for providing us with this great
opportunity to showcase our talent and skills. We would like to thank
Mr. Jignesh Sindha, PhD scholar at IIT Kharagpur from the bottom
of our hearts for his guidance and motivation throughout. Lastly, this
project would never have been completed without Mr. Sanjay,
bicycle mechanic whose technical knowledge about the little things
in a bicycle came handy throughout our journey.

\ifCLASSOPTIONcaptionsoff
  \newpage
\fi



%

%

\begin{IEEEbiography}[{\includegraphics[width=1in,height=1.25in,clip,keepaspectratio]{picture}}]{John Doe}
\blindtext
\end{IEEEbiography}




\end{document}

%% file: balancing.pdf_tex
\begingroup%
  \makeatletter%
  \providecommand\color[2][]{%
    \errmessage{(Inkscape) Color is used for the text in Inkscape, but the package 'color.sty' is not loaded}%
    \renewcommand\color[2][]{}%
  }%
  \providecommand\transparent[1]{%
    \errmessage{(Inkscape) Transparency is used (non-zero) for the text in Inkscape, but the package 'transparent.sty' is not loaded}%
    \renewcommand\transparent[1]{}%
  }%
  \providecommand\rotatebox[2]{#2}%
  \ifx\svgwidth\undefined%
    \setlength{\unitlength}{384.99999519bp}%
    \ifx\svgscale\undefined%
      \relax%
    \else%
      \setlength{\unitlength}{\unitlength * \real{\svgscale}}%
    \fi%
  \else%
    \setlength{\unitlength}{\svgwidth}%
  \fi%
  \global\let\svgwidth\undefined%
  \global\let\svgscale\undefined%
  \makeatother%
  \begin{picture}(1,1.36363634)%
    \put(0,0){\includegraphics[width=\unitlength,page=1]{balancing.pdf}}%
  \end{picture}%
\endgroup%

%% file: balancing2.pdf_tex
\begingroup%
  \makeatletter%
  \providecommand\color[2][]{%
    \errmessage{(Inkscape) Color is used for the text in Inkscape, but the package 'color.sty' is not loaded}%
    \renewcommand\color[2][]{}%
  }%
  \providecommand\transparent[1]{%
    \errmessage{(Inkscape) Transparency is used (non-zero) for the text in Inkscape, but the package 'transparent.sty' is not loaded}%
    \renewcommand\transparent[1]{}%
  }%
  \providecommand\rotatebox[2]{#2}%
  \ifx\svgwidth\undefined%
    \setlength{\unitlength}{405bp}%
    \ifx\svgscale\undefined%
      \relax%
    \else%
      \setlength{\unitlength}{\unitlength * \real{\svgscale}}%
    \fi%
  \else%
    \setlength{\unitlength}{\svgwidth}%
  \fi%
  \global\let\svgwidth\undefined%
  \global\let\svgscale\undefined%
  \makeatother%
  \begin{picture}(1,0.78518514)%
    \put(0,0){\includegraphics[width=\unitlength,page=1]{balancing2.pdf}}%
  \end{picture}%
\endgroup%

%% file: geometry.pdf_tex
\begingroup%
  \makeatletter%
  \providecommand\color[2][]{%
    \errmessage{(Inkscape) Color is used for the text in Inkscape, but the package 'color.sty' is not loaded}%
    \renewcommand\color[2][]{}%
  }%
  \providecommand\transparent[1]{%
    \errmessage{(Inkscape) Transparency is used (non-zero) for the text in Inkscape, but the package 'transparent.sty' is not loaded}%
    \renewcommand\transparent[1]{}%
  }%
  \providecommand\rotatebox[2]{#2}%
  \ifx\svgwidth\undefined%
    \setlength{\unitlength}{516.00002018bp}%
    \ifx\svgscale\undefined%
      \relax%
    \else%
      \setlength{\unitlength}{\unitlength * \real{\svgscale}}%
    \fi%
  \else%
    \setlength{\unitlength}{\svgwidth}%
  \fi%
  \global\let\svgwidth\undefined%
  \global\let\svgscale\undefined%
  \makeatother%
  \begin{picture}(1,0.86434102)%
    \put(0,0){\includegraphics[width=\unitlength,page=1]{geometry.pdf}}%
  \end{picture}%
\endgroup%

%% file: force_analysis.pdf_tex
\begingroup%
  \makeatletter%
  \providecommand\color[2][]{%
    \errmessage{(Inkscape) Color is used for the text in Inkscape, but the package 'color.sty' is not loaded}%
    \renewcommand\color[2][]{}%
  }%
  \providecommand\transparent[1]{%
    \errmessage{(Inkscape) Transparency is used (non-zero) for the text in Inkscape, but the package 'transparent.sty' is not loaded}%
    \renewcommand\transparent[1]{}%
  }%
  \providecommand\rotatebox[2]{#2}%
  \ifx\svgwidth\undefined%
    \setlength{\unitlength}{639.99998078bp}%
    \ifx\svgscale\undefined%
      \relax%
    \else%
      \setlength{\unitlength}{\unitlength * \real{\svgscale}}%
    \fi%
  \else%
    \setlength{\unitlength}{\svgwidth}%
  \fi%
  \global\let\svgwidth\undefined%
  \global\let\svgscale\undefined%
  \makeatother%
  \begin{picture}(1,0.67343754)%
    \put(0,0){\includegraphics[width=\unitlength,page=1]{force_analysis.pdf}}%
  \end{picture}%
\endgroup%

%% file: drive.pdf_tex
\begingroup%
  \makeatletter%
  \providecommand\color[2][]{%
    \errmessage{(Inkscape) Color is used for the text in Inkscape, but the package 'color.sty' is not loaded}%
    \renewcommand\color[2][]{}%
  }%
  \providecommand\transparent[1]{%
    \errmessage{(Inkscape) Transparency is used (non-zero) for the text in Inkscape, but the package 'transparent.sty' is not loaded}%
    \renewcommand\transparent[1]{}%
  }%
  \providecommand\rotatebox[2]{#2}%
  \ifx\svgwidth\undefined%
    \setlength{\unitlength}{490.00000961bp}%
    \ifx\svgscale\undefined%
      \relax%
    \else%
      \setlength{\unitlength}{\unitlength * \real{\svgscale}}%
    \fi%
  \else%
    \setlength{\unitlength}{\svgwidth}%
  \fi%
  \global\let\svgwidth\undefined%
  \global\let\svgscale\undefined%
  \makeatother%
  \begin{picture}(1,1.13673468)%
    \put(0,0){\includegraphics[width=\unitlength,page=1]{drive.pdf}}%
  \end{picture}%
\endgroup%

%% file: flow_chart.pdf_tex
\begingroup%
  \makeatletter%
  \providecommand\color[2][]{%
    \errmessage{(Inkscape) Color is used for the text in Inkscape, but the package 'color.sty' is not loaded}%
    \renewcommand\color[2][]{}%
  }%
  \providecommand\transparent[1]{%
    \errmessage{(Inkscape) Transparency is used (non-zero) for the text in Inkscape, but the package 'transparent.sty' is not loaded}%
    \renewcommand\transparent[1]{}%
  }%
  \providecommand\rotatebox[2]{#2}%
  \ifx\svgwidth\undefined%
    \setlength{\unitlength}{863.00003652bp}%
    \ifx\svgscale\undefined%
      \relax%
    \else%
      \setlength{\unitlength}{\unitlength * \real{\svgscale}}%
    \fi%
  \else%
    \setlength{\unitlength}{\svgwidth}%
  \fi%
  \global\let\svgwidth\undefined%
  \global\let\svgscale\undefined%
  \makeatother%
  \begin{picture}(1,0.58516797)%
    \put(0,0){\includegraphics[width=\unitlength,page=1]{flow_chart.pdf}}%
  \end{picture}%
\endgroup%

%% file: steering.pdf_tex
\begingroup%
  \makeatletter%
  \providecommand\color[2][]{%
    \errmessage{(Inkscape) Color is used for the text in Inkscape, but the package 'color.sty' is not loaded}%
    \renewcommand\color[2][]{}%
  }%
  \providecommand\transparent[1]{%
    \errmessage{(Inkscape) Transparency is used (non-zero) for the text in Inkscape, but the package 'transparent.sty' is not loaded}%
    \renewcommand\transparent[1]{}%
  }%
  \providecommand\rotatebox[2]{#2}%
  \ifx\svgwidth\undefined%
    \setlength{\unitlength}{846.0000346bp}%
    \ifx\svgscale\undefined%
      \relax%
    \else%
      \setlength{\unitlength}{\unitlength * \real{\svgscale}}%
    \fi%
  \else%
    \setlength{\unitlength}{\svgwidth}%
  \fi%
  \global\let\svgwidth\undefined%
  \global\let\svgscale\undefined%
  \makeatother%
  \begin{picture}(1,0.56501179)%
    \put(0,0){\includegraphics[width=\unitlength,page=1]{steering.pdf}}%
  \end{picture}%
\endgroup%

%% file: planning.pdf_tex
\begingroup%
  \makeatletter%
  \providecommand\color[2][]{%
    \errmessage{(Inkscape) Color is used for the text in Inkscape, but the package 'color.sty' is not loaded}%
    \renewcommand\color[2][]{}%
  }%
  \providecommand\transparent[1]{%
    \errmessage{(Inkscape) Transparency is used (non-zero) for the text in Inkscape, but the package 'transparent.sty' is not loaded}%
    \renewcommand\transparent[1]{}%
  }%
  \providecommand\rotatebox[2]{#2}%
  \ifx\svgwidth\undefined%
    \setlength{\unitlength}{845.00001922bp}%
    \ifx\svgscale\undefined%
      \relax%
    \else%
      \setlength{\unitlength}{\unitlength * \real{\svgscale}}%
    \fi%
  \else%
    \setlength{\unitlength}{\svgwidth}%
  \fi%
  \global\let\svgwidth\undefined%
  \global\let\svgscale\undefined%
  \makeatother%
  \begin{picture}(1,0.43431954)%
    \put(0,0){\includegraphics[width=\unitlength,page=1]{planning.pdf}}%
  \end{picture}%
\endgroup%

%% file: encoder.pdf_tex
\begingroup%
  \makeatletter%
  \providecommand\color[2][]{%
    \errmessage{(Inkscape) Color is used for the text in Inkscape, but the package 'color.sty' is not loaded}%
    \renewcommand\color[2][]{}%
  }%
  \providecommand\transparent[1]{%
    \errmessage{(Inkscape) Transparency is used (non-zero) for the text in Inkscape, but the package 'transparent.sty' is not loaded}%
    \renewcommand\transparent[1]{}%
  }%
  \providecommand\rotatebox[2]{#2}%
  \ifx\svgwidth\undefined%
    \setlength{\unitlength}{715.00000961bp}%
    \ifx\svgscale\undefined%
      \relax%
    \else%
      \setlength{\unitlength}{\unitlength * \real{\svgscale}}%
    \fi%
  \else%
    \setlength{\unitlength}{\svgwidth}%
  \fi%
  \global\let\svgwidth\undefined%
  \global\let\svgscale\undefined%
  \makeatother%
  \begin{picture}(1,0.5076923)%
    \put(0,0){\includegraphics[width=\unitlength,page=1]{encoder.pdf}}%
  \end{picture}%
\endgroup%

%% file: power.pdf_tex
\begingroup%
  \makeatletter%
  \providecommand\color[2][]{%
    \errmessage{(Inkscape) Color is used for the text in Inkscape, but the package 'color.sty' is not loaded}%
    \renewcommand\color[2][]{}%
  }%
  \providecommand\transparent[1]{%
    \errmessage{(Inkscape) Transparency is used (non-zero) for the text in Inkscape, but the package 'transparent.sty' is not loaded}%
    \renewcommand\transparent[1]{}%
  }%
  \providecommand\rotatebox[2]{#2}%
  \ifx\svgwidth\undefined%
    \setlength{\unitlength}{844.00000384bp}%
    \ifx\svgscale\undefined%
      \relax%
    \else%
      \setlength{\unitlength}{\unitlength * \real{\svgscale}}%
    \fi%
  \else%
    \setlength{\unitlength}{\svgwidth}%
  \fi%
  \global\let\svgwidth\undefined%
  \global\let\svgscale\undefined%
  \makeatother%
  \begin{picture}(1,0.61966824)%
    \put(0,0){\includegraphics[width=\unitlength,page=1]{power.pdf}}%
  \end{picture}%
\endgroup%

%% file: table_specs.pdf_tex
\begingroup%
  \makeatletter%
  \providecommand\color[2][]{%
    \errmessage{(Inkscape) Color is used for the text in Inkscape, but the package 'color.sty' is not loaded}%
    \renewcommand\color[2][]{}%
  }%
  \providecommand\transparent[1]{%
    \errmessage{(Inkscape) Transparency is used (non-zero) for the text in Inkscape, but the package 'transparent.sty' is not loaded}%
    \renewcommand\transparent[1]{}%
  }%
  \providecommand\rotatebox[2]{#2}%
  \ifx\svgwidth\undefined%
    \setlength{\unitlength}{377.00000192bp}%
    \ifx\svgscale\undefined%
      \relax%
    \else%
      \setlength{\unitlength}{\unitlength * \real{\svgscale}}%
    \fi%
  \else%
    \setlength{\unitlength}{\svgwidth}%
  \fi%
  \global\let\svgwidth\undefined%
  \global\let\svgscale\undefined%
  \makeatother%
  \begin{picture}(1,1.56233426)%
    \put(0,0){\includegraphics[width=\unitlength,page=1]{table_specs.pdf}}%
  \end{picture}%
\endgroup%

%% file: i-bike.pdf_tex
\begingroup%
  \makeatletter%
  \providecommand\color[2][]{%
    \errmessage{(Inkscape) Color is used for the text in Inkscape, but the package 'color.sty' is not loaded}%
    \renewcommand\color[2][]{}%
  }%
  \providecommand\transparent[1]{%
    \errmessage{(Inkscape) Transparency is used (non-zero) for the text in Inkscape, but the package 'transparent.sty' is not loaded}%
    \renewcommand\transparent[1]{}%
  }%
  \providecommand\rotatebox[2]{#2}%
  \ifx\svgwidth\undefined%
    \setlength{\unitlength}{721.9999875bp}%
    \ifx\svgscale\undefined%
      \relax%
    \else%
      \setlength{\unitlength}{\unitlength * \real{\svgscale}}%
    \fi%
  \else%
    \setlength{\unitlength}{\svgwidth}%
  \fi%
  \global\let\svgwidth\undefined%
  \global\let\svgscale\undefined%
  \makeatother%
  \begin{picture}(1,0.87257619)%
    \put(0,0){\includegraphics[width=\unitlength,page=1]{i-bike.pdf}}%
  \end{picture}%
\endgroup%